\documentclass[letterpaper]{article} 
\usepackage{ijcai26}  
\usepackage{times}  
\usepackage{soul}
\usepackage{url}
\usepackage[hidelinks]{hyperref}
\usepackage[utf8]{inputenc}
\usepackage[small]{caption}
\usepackage{graphicx}
\usepackage{amsmath}
\usepackage{amsthm}
\usepackage{booktabs}
\usepackage{algorithm}
\usepackage{algorithmic}
\usepackage[switch]{lineno}
\usepackage{algorithm}
\usepackage{algorithmic}
\usepackage{amsmath}
\usepackage{amssymb}
\usepackage{tikz}
\usetikzlibrary{positioning,trees}
\usepackage{xcolor}
\usepackage{colortbl}
\usepackage{subcaption}
\usepackage[margin=1in]{geometry}
\usepackage{xcolor}
\usepackage{newfloat}
\usepackage{listings}
\usepackage{booktabs}
\DeclareCaptionStyle{ruled}{labelfont=normalfont,labelsep=colon,strut=off} 
\floatstyle{ruled}
\newfloat{listing}{tb}{lst}{}
\floatname{listing}{Listing}
\author{
Liesbeth Allein$^{1,2}$
\And
Marie-Francine Moens$^1$
\affiliations
$^1$Department of Computer Science, KU Leuven, Belgium\\
$^2$Department of Electronics and Information Systems, Ghent University, Belgium\\
\emails
\{Liesbeth.Allein\}@UGent.be
}

\title{Implicit Causal Graph Construction in Text via Chain Discovery}

\begin{document}

\maketitle

\begin{abstract}
Causal graphs in text are typically populated by observable, predefined events. In contrast, we study implicit causal graph construction from text by treating each described cause–effect pair as the begin- and endpoint of an underlying latent causal graph and using large language models (LLMs) to infer intermediate causal events. We compare end‑to‑end graph construction with methods that frame the task as causal chain discovery. In the latter, graphs are built either by aggregating inferred chains or by progressively expanding partial chains through an iterative search process. We further explore ``Wisdom of the Crowd" extensions that access causal knowledge from multiple LLMs in post-hoc aggregation and collaborative inference settings. We analyze trade-offs among these approaches and evaluate the validity of inferred causal relations using a manually curated database of 1,560 scientifically validated causal pairs. This database-based evaluation is proposed as reliable, resource-efficient, and transferable to settings where ground-truth graphs are unavailable.
\end{abstract}


\section{Introduction}

When people discuss causality between two events (e.g., \textit{climate change} $\rightarrow$ \textit{drought}), they usually leave the intermediate events that explain the causal mechanisms linking the two events implicit. They assume that their conversation partners have background knowledge similar to theirs so that they infer those intermediate events themselves, or the causal relationship seems obvious to them and does not need any further explanation.

However, the causality may be entirely incorrect, merely represent correlation or even strategically used to mislead. The latter is especially troubling in topic areas such as climate change, where causality between climatic events and attribution of responsibility is central to public discussion and climate policy-making \cite{nicholson2014climate,feldman2018climate,minnerop2019climate}. Revealing the various causal mechanisms underlying events discussed in conversations and written texts step-by-step is therefore necessary to verify the validity of the causal relation attributed to these events and identify where in the reasoning process causality breaks. 

Existing work on causal graph discovery with textual events has only begun to address these challenges. Most approaches focus on retrieving explicitly stated cause-effect pairs and constructing graphs from these observable relations \cite{tan-etal-2023-recess,liu-etal-2024-identifying,vo-etal-2025-access}. Implicit causal events are left largely unexplored. Moreover, many datasets used to benchmark causal reasoning capabilities of large language models (LLMs) or to train graph construction models are automatically generated by LLMs \cite{chi2024unveiling,wang-2024-causalbench,xiong-etal-2025-com2}, thereby relying on those models’ embedded causal knowledge to define the ground truth.

In this work, we build implicit causal graphs between causal pairs extracted from text, treating the given cause as the source node and the given effect as the sink node. We compare an end-to-end graph generation approach with methods that frame graph generation as a \textit{causal chain discovery} task \cite{allein2025assessing}. This approach mirrors how people naturally reason about causal mechanisms \cite{ahn1996mechanism,walsh2011meaning}: when presented with a causal pair, they mentally construct one or more sequences of intermediary events to explain how the cause leads to the effect, e.g., \textit{climate change} $\rightarrow$ \textit{changes in precipitation patterns} $\rightarrow$ \textit{more periods of reduced rainfall} $\rightarrow$ \textit{drought}. 
Adopting ``wisdom of the crowd'' approaches, we investigate whether outputs from multiple LLMs can surface complementary causal knowledge and ultimately yield higher-quality causal graphs. 

We test our methods on two causal discovery datasets on climate change without ground-truth causal graphs. In this way, we reduce the likelihood of memorization effects and data contamination in the LLMs. Instead of residing to LLM-as-a-judge frameworks or large causal knowledge graphs for graph evaluation in such a setting, we evaluate the validity of predicted causal relations using a structured database of formulation-consistent cause-effect pairs. To build this resource, we manually extract 1,560 scientifically validated pairs from authoritative United Nations reports on climate change. 

\begin{figure*}[ht!]
    \centering
    \includegraphics[width=0.95\linewidth]{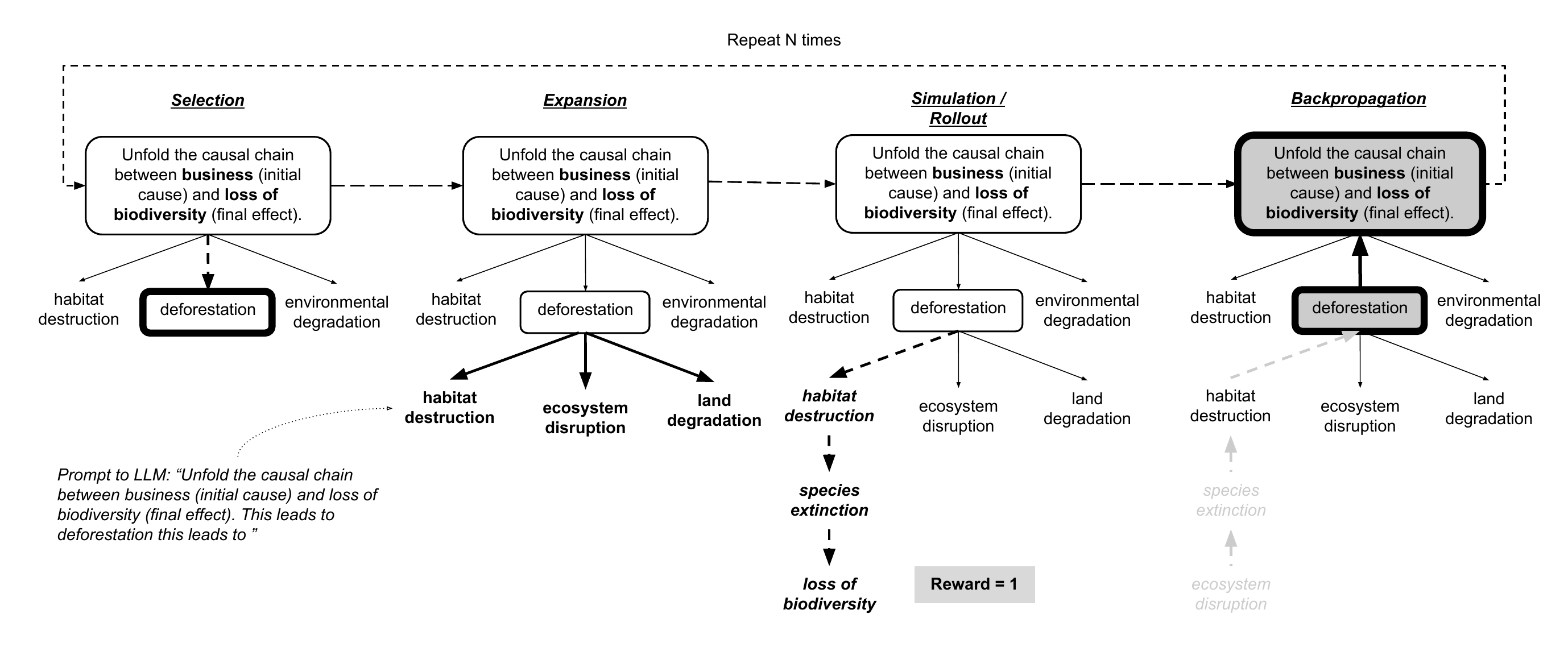}
    \caption{Overview of the Monte Carlo Tree Search (MCTS) algorithm used in graph construction via iterative causal chain discovery.}
    \label{fig:mcts_algorithm}
\end{figure*}

\paragraph{Main contributions} While recent work has started to address implicit causality (i.e., cause-effect pairs are implied rather than stated), implicit causal graph construction goes substantially further in terms of implicitness by targeting the discovery of \textit{unobserved intermediate causal mechanisms} linking events. In doing so, it shifts the focus from surface-level causal attribution to modeling underlying causal reasoning processes. We empirically compare end‑to‑end graph generation with single‑pass and iterative chain discovery methods, including multi‑LLM and collaborative chain expansion strategies. 
To support reliable evaluation without complete ground-truth graphs, we manually curate a formulation‑consistent database of scientifically validated causal pairs on climate change and present the first systematic evaluation of implicit causal graph validity in this setting.

\section{Related Work}

Prior work typically assumes access to a fixed set of candidate events or operates over events that are explicitly present in the text. Local prompting approaches construct causal graphs by evaluating individual relations and aggregating them into a joint graph. Methods rely on edge-wise prompts, where LLMs determine causality between pairs of events \cite{kiciman2023causal,long2023causal,antonucci2023zero} or triplets of events \cite{vashishtha2023causal}. They either do this by using the causal knowledge embedded in their parameters or through Retrieval-Augmented Generation, in which causality judgment is based on supporting or refuting evidence \cite{feng-etal-2025-iris}. Iterative graph-building approaches generate causal graphs step by step. \cite{jiralerspong2024efficient} applied a breadth-first approach using a complete predefined variable set. \cite{liu-etal-2024-identifying} iteratively updated event‑causality graphs extracted from verb‑phrase events mentioned in the text. Finally, multi-LLM collaborative approaches distribute causal reasoning across multiple LLMs. In \cite{koupaee-etal-2025-causal}, these models act as an expert for assessing a specific aspect of causality between two events, i.e., temporal relation, discourse relation, conditional relation, and commonsense relation, with a separate LLM resolving disagreements. By contrast, our setting specifies only the source and sink event, and requires the model to discover implicit intermediate causal events..

\section{Task Formalization}

\paragraph{Input} Given a \textit{cause-effect pair} \( E_c \rightarrow E_f \), cause \( E_c \) and effect \( E_f \) are noun phrases that describe an event or action represented as a sequence of natural language tokens in English (e.g., \textit{``greenhouse gas emissions''}). The arrow \( \rightarrow \) denotes the \textit{directionality} of causality from cause to effect. 

\paragraph{Output} The aim is to obtain a causal graph $G$, which we consider a directed acyclic graph (DAG) of at least three nodes consistent with the DAG structure proposed by \cite{pearl2009causality}. The nodes $n_i \in G$ take textual events or actions $E$ as text values; e.g., \textit{``increase in cooling aerosols''}. Events are formatted as noun phrases, 
such that formatting of the nodes' text value is consistent throughout $G$. 
The edges in $G$ encode causal relations, with their directionality indicating the flow from cause to effect.
In the implicit causal graphs, the node representing $E_c$, i.e., $n_0$, should have only outgoing edges, while the node representing $E_f$, i.e., $n_f$, should only have incoming edges. All other nodes can have both incoming and outgoing edges.

\section{Methodology} 

We investigate three strategies for implicit causal graph construction: an end-to-end approach that directly produces the graph; a two-stage approach that produces causal chains and aggregates these to form the graph; and an iterative approach where causal chain discovery progressively guides graph construction. Note that we do not intend to establish causality in the graph construction frameworks in the strict scientific sense defined by causal inference theory. Instead, we target explanatory causal reasoning as triggered by text. The task is designed to surface implicit intermediate mechanisms that people invoke when explaining why one event leads to another, rather than to validate those mechanisms through observational or interventional evidence during generation. In this respect, the causal graphs constitute structured representations of belief‑based causal reasoning pathways, which may reflect incomplete, contested, or even falsified explanations.

\subsection{End-to-End Graph Construction} \label{sec:method-endtoend}

In this baseline setting, the mapping from a given causal pair $E_c \rightarrow E_f$ to $G$ is done in a single generative process: $G = \phi_{graph}(E_c, E_f)$. Function $\phi_{graph}$ is parameterized by an LLM that takes a prompt specifying $E_c$ and $E_f$ as begin- and endpoint and produces an adjacency list of directed event pairs for constructing $G$. All prompts are included in the Appendix.

\subsection{Graph Construction via \textit{Single-Pass} Causal Chain Discovery} \label{sec:method-singlepass}

\textit{Implicit causal chain discovery} involves the inference of chain-like structures that explain the underlying causal mechanisms between two given events. 
Since human causal reasoning is likewise mechanistic, we investigate whether framing graph construction as a causal chain discovery task followed by aggregation yields more informative causal graphs. Unlike end‑to‑end graph construction, which requires the model to reason globally over graph structure, this approach decomposes graph construction into local mechanistic explanations that are subsequently composed.

We build on the approach in \cite{allein2025assessing}, where a set of $S$ intermediate causal chains linking $E_c$ to $E_f$ is inferred in a single inference pass, yielding $C = \{C_1,...,C_S\}$. Each chain $C_s$ is expressed as an ordered set of directed cause-effect pairs. 
$C$ is inferred using function $\phi_{chain}(E_c,E_f)$, which is parameterized by an LLM prompted with $E_c$ and $E_f$. The model is instructed to predict all plausible causal chains linking $E_c$ to $E_f$. Since $S$ is input-dependent, the model indirectly infers the appropriate value for $S$. 

We extend this approach by defining an aggregation function that maps $C$ to a unified causal graph: $G = \text{agg}(C)$, with $\text{agg}(\cdot)$ combining all cause-effect pairs across the predicted chains, i.e. $\bigcup_{s=1}^{S} C_s$. This aggregation strategy merges shared edges. 
As a result, $G$ emerges from the composition of independently inferred mechanistic explanations rather than from end‑to‑end graph construction.

\subsection{Graph Construction via \textit{Iterative} Causal Chain Discovery} \label{sec:method-iterative}
Single‑pass generation infers all causal structures, i.e., a graph or all chains, in one forward inference. A model has to simultaneously identify relevant intermediate events and assemble them into a coherent ordering while connecting $E_c$ to $E_f$. Here, we formulate causal graph construction as a process of successive local extensions, in which $G$ is built \textit{incrementally} from $E_c$, represented in root node $n_o$, to $E_f$, marking a terminal or leaf node $n_f$. The iterative graph construction approach therefore trades in one‑step global prediction for incremental causal inference guided by a structured search, such that alternative causal mechanisms can be systematically explored while progressively shaping the global structure of $G$. 

We operationalize this idea using Monte Carlo Tree Search (MCTS; see Figure~\ref{fig:mcts_algorithm}), which provides a framework for navigating the space of partial causal explanations. At a high level, MCTS maintains and expands a search tree whose nodes correspond to partial causal paths originating from $E_c$, each terminating at an intermediate event. The search process iteratively selects promising partial paths for expansion, generates candidate causal extensions using a next‑step inference function $\phi_{step}$ parameterized by an LLM, and evaluates their potential to contribute to a valid causal connection between $E_c$ and $E_f$. 

\subsubsection{STEP 1 -- Selection} 

The selection step identifies a path through the search tree that terminates at the next node to be expanded. Starting from $n_0$, we traverse the tree by recursively selecting child nodes using the Upper Confidence Bound for Trees (UCT) criterion \cite{kocsis2006bandit}. At each depth $i$, a child node $n_{i+1} \in Children(n_i)$ is selected according to
\begin{equation}
n_{i+1} = \arg \max_{n \in Children(n_i)} \frac{Q(n)}{V(n)} + \lambda \cdot \sqrt{\frac{\ln V(n_i)}{V(n)}}
\end{equation}
with $Q(n)$ the cumulative reward of $n$, $V(n)$ the number of visits to $n$, $V(n_i)$ the number of visits to $n_i$, and $\lambda = 2.0$ the constant exploration weight. Each node $n$ represents a text-valued partial state in the search process, i.e., an event in a causal graph, and maintains its associated cumulative reward and visit count, which are updated as the tree is traversed and expanded. The selection process terminates at depth $d$ when the selected node $n_d$ is either terminal, i.e., represents $E_f$, or expandable. We define an expandable node as a non-terminal node that can be further extended by at least one child node. Thus, although nodes may exist at depth $d$ in the tree, $n_d$ refers to the node reached by following UCT from $n_0$, and it is the earliest node along the selected path $Path = (n_0, n_1, \dots, n_d)$ that is expandable.

\subsubsection{STEP 2 -- Expansion} 
In the expansion step, $J = 3$ nodes are added as child nodes of $n_d$, such that $Children(n_d) = \{n^{(1)}_{d+1},n^{(2)}_{d+1},n^{(3)}_{d+1}\}$. Each child node $n^{(j)}_{d+1}$ is generated by sampling an event string $a^{(j)}$ from $\phi_{step}$ conditioned on $Path$:
\begin{equation}
    a^{(j)} \sim \phi_{step}(\cdot|s(Path))
\end{equation}
with function $s(\cdot)$ transforming $Path$ into a string by concatenating the event text values of its nodes and separating them using guidance prompt $g =$ \textit{``this leads to''}; $s(Path) = n_0;g;n_1;g;...;n_d;g$. Each $a^{(j)} = (x_1, x_2, ..., x_T)$ is a sequence of $T$ tokens that represents an event, where each token $x_t$ is generated autoregressively:
\begin{equation}
    x_{t+1} \sim \text{Sample}(\hat{\mathbf{y}}_t)
\end{equation}
\begin{equation}
    \hat{\mathbf{y}}_t = \text{softmax}(\mathbf{l}_t) 
\quad \Rightarrow \quad 
\hat{y}_{t,i} = \frac{e^{l_{t,i}}}{\sum_{v=1}^{V} e^{l_{t,v}}}
\end{equation}
\begin{equation}
    \mathbf{l}_t = \mathbf{z}_t \cdot \mathbf{W}+\mathbf{b}
\end{equation}
\begin{equation}
    \mathbf{z}_t = \psi(s(Path),x_{<t})
\end{equation}
The next token $x_{t+1}$ is sampled from $\hat{\mathbf{y}}_t \in \mathbb{R}^V$, which presents the probability distribution over vocabulary $V$. $\hat{\mathbf{y}}_t$ is obtained by applying a softmax function over the logits vector $\mathbf{l}_t$, which in turn is obtained by projecting the final hidden state vector $\mathbf{z}_t$ produced by the Transformer layers $\psi$ of the LLM parameterizing $\phi_{step}$ into the vocabulary space using a linear transformation. $\psi$ takes the sequence of tokens generated up to $t$, i.e., $x_{<t}$, and $s(Path)$. Generation of $a^{(j)}$ terminates when an end-of-sentence token is emitted or when a maximum token length is reached. $T$ is therefore variable. Finally, we assign $a^{(j)}$ as the text value of $n^{(j)}_{d+1}$.

\paragraph{Promote diversity among children} We encourage diversity among $Children(n_d)$ by \emph{reducing lexical overlap} between the child nodes.   
When generating $n_{d+1}^{(j)}$, $D^{(j)}$ is the set of unique tokens used in the representation of the other child nodes already produced in $Children(n_d)$: $D^{(j)} = \bigcup_{i=1}^{j-1} \text{set}(n_{d+1}^{(i)})$, with $\text{set}(n_{d+1}^{(i)}) = \{ x_t \mid 1 \leq t \leq T \}$ as the set of unique tokens in the text value of $n_{d+1}^{(i)}$. We give $D^{(j)}$ as additional input to $\phi_{step}$ such that $a^{(j)} \sim \phi_{step}(\cdot|s(Path),D^{(j)})$, and modify logits vector $\mathbf{l}_t$ by overwriting the logit of all tokens in $D^{(j)}$ with a large negative value (e.g., $-100$):
\begin{equation}
    l_{t,v} = \begin{cases}
-100 & \text{if } x_v \in D^{(j)} \\
l_{t,v} & \text{otherwise}
\end{cases}
\end{equation}
with $l_{t,v}$ the logits of token $x_v$ at position $v$ in $V$. The probability of the tokens in $D^{(j)}$ will be close to zero after applying the softmax function since $e^{-100} \approx 0$. As a result, it will not be considered during sampling. While this approach operates at the token level, it reduces near-duplicate outputs (e.g., \textit{deforestation} vs. \textit{deforesting}). 

\paragraph{Avoid within-chain cycles} In an effort to avoid cycles in the chain and constrain outputs to DAG structures consistent with Pearl’s causal framework \cite{pearl2009causality}, we further suppress tokens that appear in the current path $Path$ to avoid within-chain cycles. Let $B$ be the set of unique tokens composing the events in \textit{Path}: $B = \bigcup_{i=1}^{d} \text{set}(n_i)$, with $\text{set}(n_i) = \{ x_t \mid 1 \leq t \leq T \}$. 
We give $B$ as additional input to $\phi_{step}$; $a^{(j)} \sim \phi_{step}(\cdot|Path,D^{(j)},B)$ and update the logits vector $\mathbf{l}$:
\begin{equation}
    l_{t,v} = \begin{cases}
-100 & \text{if } x_v \in D^{(j)} \cup B \\
l_{t,v} & \text{otherwise}
\end{cases}
\end{equation}

While this method is effective at the lexical level, it may not capture semantically equivalent events. 

\paragraph{Capture recurring events} A candidate child node can point to a similar event captured in any other node in $G$. To capture similar events, we \emph{assess lexical similarity} between the candidate child and each $n$ in $G$ through fuzzy matching.
More specifically, we compute the \textit{normalized Indel similarity}; $\forall n  \in G$:

\begin{equation} \label{eq:fuzzymatching}
\text{fuzzy}(a^{(j)}, n) = 1 - \text{norm\_distance}(a^{(j)}, n)
\end{equation}
\begin{equation}
\text{norm\_distance}(a^{(j)}, n) = \frac{\text{distance}(a^{(j)}, n)}{\text{len}(a^{(j)}) + \text{len}(n)}
\end{equation}
\begin{equation}
\text{distance}(a^{(j)}, n) = \text{len}(a^{(j)}) + \text{len}(n) - 2 \times \text{lcs\_sim}(a^{(j)}, n)
\label{eq:distance}
\end{equation}

where $\text{len}(n)$ denotes the number of characters in the text value of $n$, and lcs\_sim$(\cdot)$ is the length of the longest common subsequence between two strings. Equation \ref{eq:distance} captures the minimum number of \textit{insertions and deletions} required to transform one sequence into another.

If the similarity between the event in the candidate child and its best match (= the node with a text value to which the candidate child is most similar, i.e., the highest similarity value) exceeds a certain threshold (here, $ \geq.85$ which is a rather strict threshold), the best match is considered the child, i.e., $n^{(j)}_{d+1} = n$. Otherwise, $n^{(j)}_{d+1} = a^{(j)}$.
Note that in the first case, the number of visits and rewards to the node, $V(n)$ and $Q(n)$ respectively, are preserved.

While fuzzy matching is an efficient and time-inexpensive method for capturing lexical similarity, it does not guarantee semantic equivalence and it misses paraphrases (e.g., \textit{heavy rainfall} vs. \textit{intense precipitation}). These would be captured better with embedding-based approaches relying on cosine similarity. However, these require more computation power and time, which would substantially slow down the graph building process when $G$ becomes larger.

\subsubsection{STEP 3 -- Rollout}

Once all $J$ children are added during expansion, we simulate a complete causal chain from $n^{(j)}_{d+1}$ to a terminal state using $\phi_{step}$.
The rollout continues until a terminal condition is met, i.e., a maximum rollout length ($=8$) or $E_f$ is reached (i.e., $a =$ \textit{``end of chain: $E_f$''}). The resulting sequence is evaluated using a straightforward reward function $R$, which assigns a reward $r \in [0, 1]$: $r=1$ if the final effect is reached; $r=0$, otherwise.

The reward function could be arguably extended with intermediate rewards that score the scientific correctness of each transition. However, this would require a closed and exhaustive ontology of causal relations, which is unrealistic in an open-world setting. Alternatives such as LLM-as-a-judge or knowledge-graph-based rewards are unreliable due to unstable causal reasoning and the lack of guarantees about the scientific validity of encoded causal relations. Since our goal is to train in a realistic setting with limited ground truth, we keep $R$ as such.

The generated nodes are included in the rollout in $G$ to accelerate causal graph building. This means that we should have at least one full causal chain very early in the causal graph construction process, possibly as early as after the first iteration. Here, we also capture events that are similar to events contained in another node already contained in the graph through fuzzy matching (see \textit{capture recurring events} in STEP 2).

\subsubsection{STEP 4 -- Backpropagation}

The reward $r$ obtained from the rollout is propagated back through the nodes in $Path$. 
For each node $n \in Path$, we update:
\begin{equation}
    V(n) \leftarrow V(n) + 1
\end{equation}
\begin{equation}
    Q(n) \leftarrow Q(n) + r
\end{equation}

\subsection{Single Model versus Wisdom of the Crowd}

All graph construction methods are evaluated under single-model and Wisdom of the Crowd (WoC) instantiations. Through WoC, we access the causal knowledge encoded in multiple LLMs for constructing the graph. \textbf{[Single Model]} $\phi_{graph}$, $\phi_{chain}$, and $\phi_{step}$ are instantiated by one LLM, such that the causal events for constructing $G$ originate from a single LLM. \textbf{[Post-hoc WoC]} A method is run $N$ times, where $\phi_{graph}$, $\phi_{chain}$, or $\phi_{step}$ is instantiated by one LLM from a set of $N$ LLMs. Afterwards, all $N \in \{3,9\}$ generated causal structures are aggregated into a joint causal graph. \textbf{[Collaborative WoC]} At each iteration, $\phi_{step}$ is instantiated by randomly selecting a LLM from a predefined set of LLMs.

\section{Experimental Setup}

\subsection{Data} We use cause-effect pairs from PolarIs3CAUS \cite{pineda2025polaris3} and PolarIs4CAUS \cite{pineda2025polaris4}. In both datasets, argumentation experts manually extracted English causal events from climate change discussions sourced from Reddit (PolarIs3CAUS, 95 pairs) and X (PolarIs4CAUS, 181 pairs). 
Although these datasets are moderate in size and focused on a specific topic, they are particularly well suited to our task: all causal events have been reformulated as noun phrases, aligning with the targeted event representation in our graph formulation. Moreover, the data is recent enough to extend beyond the training cutoff dates of current LLMs and the datasets do not provide ground-truth graphs, reducing the likelihood of memorization effects and data contamination.

\subsection{Large Language Models} 

For end-to-end graph construction: GPT4o‑mini, Falcon3‑7b, and Qwen2‑7b. For graph construction via single‑pass causal chain discovery: o1, o1‑mini, GPT4o, DeepSeek R1, Llama 3.1 Nemotron, Llama 3 70b, Mistral Nemo, Mixtral, and Phi 4‑mini. Finally, for graph construction via iterative causal chain discovery: GPT4o‑mini, Falcon3‑7b, and Qwen2‑7b. 

\section{Evaluation: Structural Properties} \label{sec:evaluation_structural_properties}

\paragraph{Metrics} \textit{Graph density} measures how many edges exist compared to the maximum possible in a DAG, with high density indicating higher level of interconnection. \textit{Longest path length} indicates the depth of dependency chains, with a longer path implying more detailed causal chains.
\textit{Average node connectivity} indicates the average connectivity of a graph, which is the average local node connectivity over all pairs of nodes in $G$.

\textit{Cycle ratio} indicates the ratio of graphs that contain a cycle. While density captures edge abundance, longest path length reflects dependency depth, average node connectivity indicates structural robustness, and cycle ratio highlights the presence of cycles that break acyclicity.

\begin{table}[]
    \centering
    \tiny
    \setlength{\tabcolsep}{1pt}
    \begin{tabular}{p{1.7cm}p{0.8cm}p{0.8cm}p{0.8cm}p{0.4cm}|p{0.8cm}p{0.8cm}p{0.8cm}p{0.4cm}}
        \toprule
        & \multicolumn{4}{c}{PolarIs3CAUS} & \multicolumn{4}{c}{PolarIs4CAUS} \\
        \midrule
        \textbf{Setup} 
        & \textbf{Graph Density ($\uparrow$)} 
        & \textbf{Longest Path Length ($\uparrow$)} 
        & \textbf{Avg. Node Connectivity ($\uparrow$)} 
        & \textbf{Cycle Ratio ($\downarrow$)} 
        & \textbf{Graph Density ($\uparrow$)} 
        & \textbf{Longest Path Length ($\uparrow$)} 
        & \textbf{Avg. Node Connectivity ($\uparrow$)} 
        & \textbf{Cycle Ratio ($\downarrow$)} \\
        \midrule
        \multicolumn{9}{c}{\textit{\textbf{End-to-End} Graph Construction}} \\
        \midrule
        \textit{Single Model}* 
        & \textbf{0.29}\textsubscript{±0.11} & \textit{3.49}\textsubscript{±1.78} & \underline{0.42}\textsubscript{±0.11} & \textbf{0.01}
        & \textbf{0.29}\textsubscript{±0.11} & \textit{3.53}\textsubscript{±1.51} & \textbf{0.44}\textsubscript{±0.10} & \textbf{0.02}  \\
        \rowcolor{gray!20}\textit{Post-hoc WoC} 
        & 0.16\textsubscript{±0.17} & 5.18\textsubscript{±3.23} & 0.29\textsubscript{±0.16} & \underline{0.02} 
        & 0.14\textsubscript{±0.16} & 5.55\textsubscript{±2.80} & 0.27\textsubscript{±0.14} & \underline{0.05} \\
        
        \midrule
        \multicolumn{9}{c}{\textit{Via \textbf{Single-Pass} Causal Chain Discovery}} \\
        \midrule

        \textit{Single Model}** 
        & 0.07\textsubscript{±0.04} & 5.69\textsubscript{±1.88} & 0.21\textsubscript{±0.09} & \underline{0.02}
        & 0.07\textsubscript{±0.03} & 5.83\textsubscript{±2.08} & 0.21\textsubscript{±0.08} & \textbf{0.02} \\

        \rowcolor{gray!20}\textit{Post-hoc WoC}*
        & 0.03\textsubscript{±0.01} & 7.05\textsubscript{±2.40} & 0.08\textsubscript{±0.03} & 0.08
        & 0.03\textsubscript{±0.01} & 7.23\textsubscript{±2.97} & 0.09\textsubscript{±0.04} & 0.07 \\

        \rowcolor{gray!20}\textit{Post-hoc WoC}**
        & \textit{0.01}\textsubscript{±0.00} & \textbf{9.29}\textsubscript{±2.88} & \textit{0.03}\textsubscript{±0.01} & \textit{0.24}
        & \textit{0.01}\textsubscript{±0.00} & \textbf{10.41}\textsubscript{±4.04} & \textit{0.03}\textsubscript{±0.02} & \textit{0.30} \\

        \midrule
        \multicolumn{9}{c}{\textit{Via \textbf{Iterative} Causal Chain Discovery}} \\
        \midrule

        \textit{Single Model} (\textit{10})* 
        & \underline{0.26}\textsubscript{±0.10} & 4.06\textsubscript{±1.55} & \textbf{0.43}\textsubscript{±0.03} & -
        & \underline{0.25}\textsubscript{±0.10} & 4.22\textsubscript{±1.56} & \textbf{0.44}\textsubscript{±0.04} & - \\

        \textit{Single Model} (\textit{20})* 
        & 0.25\textsubscript{±0.10} & 4.69\textsubscript{±1.98} & \textbf{0.43}\textsubscript{±0.04} & -
        & \underline{0.25}\textsubscript{±0.10} & 4.52\textsubscript{±1.70} & \textbf{0.44}\textsubscript{±0.03} & - \\

        \rowcolor{gray!20}\textit{Post-hoc WoC} (\textit{10}) 
        & 0.10\textsubscript{±0.08} & 6.96\textsubscript{±2.02} & 0.30\textsubscript{±0.11} & -
        & 0.11\textsubscript{±0.11} & 6.96\textsubscript{±2.35} & \underline{0.32}\textsubscript{±0.12} & - \\

        \rowcolor{gray!20}\textit{Post-hoc WoC} (\textit{20}) 
        & 0.08\textsubscript{±0.08} & \underline{8.62}\textsubscript{±3.24} & 0.29\textsubscript{±0.10} & -
        & 0.10\textsubscript{±0.10} & \underline{7.73}\textsubscript{±2.78} & 0.31\textsubscript{±0.11} & - \\

        \textit{Collab. WoC} (\textit{10})  
        & 0.13\textsubscript{±0.10} & 5.53\textsubscript{±2.77} & 0.27\textsubscript{±0.12} & -
        & 0.13\textsubscript{±0.09} & 5.48\textsubscript{±2.69} & 0.29\textsubscript{±0.10} & - \\

        \textit{Collab. WoC} (\textit{20}) 
        & 0.19\textsubscript{±0.13} & 4.07\textsubscript{±3.51} & 0.29\textsubscript{±0.11} & -
        & 0.14\textsubscript{±0.09} & 5.18\textsubscript{±2.84} & 0.27\textsubscript{±0.10} & - \\

        \bottomrule
    \end{tabular}
    \caption{Evaluation of \textbf{structural properties} of the generated causal graphs for PolarIs3CAUS and PolarIs4CAUS across graph construction methods. Reported values denote the mean and standard deviation (±) computed over all predicted graphs. * indicates averages over three models or model combinations, while ** indicates averages over nine. For iterative methods, results are shown for 10 and 20 MCTS iterations. Best-performing results are in \textbf{bold}, second-best are \underline{underlined}, and worst-performing results are in \textit{italics}.}
    \label{tab:evaluation_structural_properties}
\end{table}

\paragraph{Results} 
We report mean performance results in Table \ref{tab:evaluation_structural_properties} and include per-LLM results in the Appendix. Cycle ratio results are not reported for evaluation of the iterative causal chain discovery, as the method is defined to avoid cycles. Graph construction methods seem to strongly affect structural characteristics. Graph construction via single-pass causal chain discovery produces the sparsest graphs with low node connectivity but tends to produce longer paths. The other two approaches yield denser graphs, reflecting more exploratory linking, with shortest path lengths for end-to-end graph construction. In post-hoc and collaborative WoC settings, cycles occur more frequently as a result of the aggregation procedure, which is also seen in these settings with the iterative causal chain discovery method.

\paragraph{Discussion} Our findings demonstrate trade-offs in terms of control in the generation process, structural density and informativeness of the produced graphs, and stability in causal reasoning performance. The proposed iterative method has clear advantages as it promotes diversity in the generated chains and avoids within-chain cycles in a convenient way. Graph construction is complicated by event matching. Differences in textual phrasing and specificity between events can cause near-misses that are semantically aligned but fail strict matching. Embedding-based semantic similarity alone tends to be too coarse for these fine-grained distinctions, particularly in cases where spatiotemporal qualifiers matter.

\section{Evaluation: Causal Validity}

\subsection{Ground-Truth Reference Database}

Since ground-truth causal graphs are unavailable, we construct a database of scientifically validated causal relations on climate change to assess the validity of the predicted graphs. Existing knowledge graphs that focus exclusively on causal relations \cite{heindorf2020causenet,9706608,10.5555/3491440.3491942}
or embed them alongside other types of knowledge \cite{speer2017conceptnet,sap2019atomic,hwang2021comet} are unreliable for this purpose. These resources are typically built automatically from a large collection of web-crawled texts using pattern-matching approaches, which does not guarantee the scientific validity of all encoded causal relations. LLM-as-a-judge frameworks are likewise unreliable due to unstable causal reasoning capabilities, e.g., high sensitivity to prompt formulations, reliance on position heuristics, and inability to distinguish between causality reported by truthful versus untruthful sources in the training data.

\paragraph{Data source} We source science-for-policy reports on climate change from the Intergovernmental Panel on Climate Change (IPCC)
, which is the United Nations body responsible for assessing the science related to climate change, including its causes, impacts, future risks, and potential strategies for adaptation and mitigation. Their periodical reports synthesize the current state of scientific knowledge. We prefer these reports over scientific papers as they span a broader range of aspects involved in climate change (e.g., biological, physical, and social) and typically discuss a larger and more diverse set of causal relations within a single document. 

\paragraph{Data annotation} We let two experts annotators manually extract 653 unique cause-effect pairs from 170 statements sampled from the IPCC reports \cite{allein2026climatecause}. The events are manually normalized to noun phrase, as done in PolarIs3CAUS and PolarIs4CAUS, to facilitate event matching during evaluation. Our goal is not to construct an exhaustive database of climate-related causal relations from these reports, but rather to demonstrate how a curated, scientifically grounded causal database can be used to meaningfully evaluate causal graphs.

\paragraph{Database construction} 
Starting from the set of 653 cause-effect pairs, denoted as $R^*$, we extract all unique events to form reference event set $S^*$, with $|S^*| = 435$. We then construct a dictionary $D^*$ that maps each event $E_s \in S^*$ to an integer index via $id(E_s)$. From $R^*$, we build a directed graph $G^*$, where each cause event is connected to its corresponding effect event by a directed edge. $G^*$ has 435 nodes and 653 edges. From $G^*$, we derive a binary reachability matrix $M^* \in \{0,1\}^{|S^*| \times |S^*|}$, where an entry is set to $1$ if there exists a (possibly multi-hop) directed path from $E_{i}$ to $E_{j}$ in $G^*$, and $0$ otherwise:

\begin{equation}
M^*(id(E_{i}), id(E_{j})) =
\begin{cases}
1 & \begin{aligned}
    &\text{if a path exists from}\\
    & E_{i} \text{ to } E_{j} \text{ in } G^*
    \end{aligned} \\
0 & \text{otherwise}
\end{cases}
\end{equation}

By including transitive reachability, $M^*$ captures not only explicitly annotated causal relations but also indirect causal dependencies implied by $G^*$.
$M^*$ contains 1,560 reachable event pairs that define the set of potentially verifiable causal relations in the database. We use $M^*$ as a look-up matrix to determine whether a predicted event pair is scientifically valid.

\subsection{Methods for Event Matching and Causal Pair Validation}

\subsubsection{Event Matching} \label{sec:event_matching}

Since an event $E$ in a generated graph may not appear verbatim in $S^*$, we identify proxy event set $E^{*}$, which corresponds to the set of most similar events to $E$ in $S^*$ retrieved via fuzzy matching; $\text{fuzzy}(\cdot)$ (Equation \ref{eq:fuzzymatching}). Specifically, we retain all events for which similarity exceeds a threshold $\tau$:
\begin{equation} \label{eq:event_matching}
    E^{*} = \{E_s \in S^* \mid \text{fuzzy}(E, E_s) \geq \tau\}
\end{equation}
Retaining all proxy events above $\tau$, rather than selecting only the closest match, makes the evaluation more robust to paraphrasing and allows semantically equivalent events that participate in different parts of $G^*$ to be considered. We set $\tau = .75$, which we found to balance semantic fidelity and coverage. While embedding-based semantic similarity is an appropriate alternative, we opt for fuzzy matching to maintain consistency with how similarity is computed in our graph construction approach via iterative causal chain discovery.

\subsubsection{Causal Pair Validation} \label{sec:causal_pair_validation}

Given the respective proxy event sets $E_x^*$ and $E_y^*$ of a given causal pair $E_x \rightarrow E_y$ (Equation~\ref{eq:event_matching}), we evaluate the validity of the causal relation between $E_x$ and $E_y$ as follows:

The relation is \textbf{valid} if there exists $E_i \in E_{x}^{*}$ and $E_j \in E_{y}^{*}$ such that $M^*(id(E_i),id(E_j)) = 1$, and if there does not exist $E_i \in E_{x}^{*}$ and $E_j \in E_{y}^{*}$ such that $M^*(id(E_j),id(E_i)) = 1$, indicating support for the reversed relation;

The relation is \textbf{invalid} if there exists $E_i \in E_{x}^{*}$ and $E_j \in E_{y}^{*}$ such that $M^*(id(E_j),id(E_i)) = 1$;

The relation is \textbf{not verifiable} using our reference database if either (i) no proxy event exceeds $\tau$ for $E_x$ or $E_y$, i.e., $E_x^{*} = \emptyset$ or $E_y^{*} = \emptyset$, or (ii) no path exists between any proxy events in either direction, i.e., for all $E_i \in E_{x}^{*}$ and $E_j \in E_{y}^{*}$, both $M^*(id(E_i), id(E_j)) = 0$ and $M^*(id(E_j), id(E_i)) = 0$.

\begin{table}[t]
    \centering
    \tiny
    \setlength{\tabcolsep}{1pt}
    \begin{tabular}{p{1.7cm}p{0.6cm}p{0.6cm}p{1cm}p{0.6cm}|p{0.6cm}p{0.6cm}p{1cm}p{0.6cm}}
        \toprule
        & \multicolumn{4}{c}{\textbf{PolarIs3CAUS}} & \multicolumn{4}{c}{\textbf{PolarIs4CAUS}} \\
        \midrule
        \textbf{Setup} 
        & \textbf{Total \# Pairs} 
        & \textbf{with CE match (\%)}
        & \textbf{+ GT path (\%)}
        & \textbf{Acc}
        & \textbf{Total \# Pairs} 
        & \textbf{with CE match (\%)}
        & \textbf{+ GT path (\%)}
        & \textbf{Acc} \\
        \midrule
        \multicolumn{9}{c}{\textit{\textbf{End-to-End} Graph Construction}} \\
        \midrule

        \textit{Single Model} 
        & \textit{981} & .0499 & .1020 (5) & \textbf{1.0000}
        & \textit{1,744} & .0728 & \underline{.1575 (20)} & \textbf{1.0000} \\

        \rowcolor{gray!30}\textit{Post-hoc WoC}
        & 1,423 & .0478 & .1029 (7) & .8571
        & \textit{1,423} & .0478 & .1029 (7) & .8571 \\

        \midrule
        \multicolumn{9}{c}{\textit{Via \textbf{Single-Pass} Causal Chain Discovery}} \\
        \midrule

        \textit{Single Model} 
        & 2,900 & \textit{.0238} & \textit{.0870 (6)} & \textbf{1.0000}
        & 5,606 & .0474 & \textit{.0865 (23)} & \textbf{1.0000} \\

        \rowcolor{gray!30}\textit{Post-hoc WoC*}
        & \underline{7,978} & .0440 & \textbf{.1994 (70)} & \textbf{1.0000}
        & \underline{20,442} & .0465 & \textbf{.1903 (181)} & \underline{.9670} \\

        \rowcolor{gray!30}\textit{Post-hoc WoC**}
        & \textbf{23,861} & .0369 & \underline{.1602 (141)} & \underline{.9929}
        & \textbf{76,812} & .0496 & .1432 (545) & .9523 \\

        \midrule
        \multicolumn{9}{c}{\textit{Via \textbf{Iterative} Causal Chain Discovery}} \\
        \midrule

        \textit{Single Model (10)} 
        & 1,999 & \textbf{.0735} & .0884 (13) & .9231
        & 4,521 & \underline{.0807} & .1534 (56) & .8393 \\

        \textit{Single Model (20)}
        & 3,734 & .0640 & .0879 (21) & \textit{.5238}
        & 7,459 & \textbf{.0815} & .1447 (88) & \textit{.8295} \\

        \rowcolor{gray!30}\textit{Post-hoc WoC (10)}
        & 2,184 & \underline{.0687} & .0933 (14) & .9286
        & 4,932 & .0783 & .1528 (59) & .8475 \\

        \rowcolor{gray!30}\textit{Post-hoc WoC (20)}
        & 3,924 & .0617 & .0909 (22) & .5455
        & 7,881 & .0799 & .1397 (88) & .8295 \\

        \textit{Collab. WoC (10)}
        & 1,025 & .0478 & .1429 (7) & .8571
        & 2,602 & \textit{.0411} & .1402 (15) & .8667 \\

        \textit{Collab. WoC (20)}
        & 1,704 & .0399 & .0882 (6) & .6667
        & 2,696 & .0579 & .1410 (22) & .8636 \\

        \bottomrule
    \end{tabular}
     \caption{Evaluation of \textbf{validity} of the causal graphs generated by GPT4o. Total number of causal pairs that are included in the generated graphs (\textit{Total \# Pairs}), with the proportion of pairs that have a proxy events for both cause and effect in the reference database (\textit{with CE match}) and the proportion of these pairs that have a directed path in the reference database (\textit{+ GT path}) with absolute pair count in brackets. The accuracy score reflects the validity of the latter pairs (\textit{Acc}). * and ** denote the aggregation of three and nine LLMs.}
    \label{tab:evaluation_validity}
\end{table}

\subsection{Evaluation}

Our evaluation assesses whether a causal relation $E_i \rightarrow E_j$ in a generated graph $G$ is supported by the reference database. This assessment builds directly on the methods for event matching (§\ref{sec:event_matching}) and causal pair validation (§\ref{sec:causal_pair_validation}). We first retrieve the proxy event set \(E_i^* \subseteq S^*\) and \(E_j^* \subseteq S^*\) via fuzzy matching (Equation~\ref{eq:event_matching}). Causal pairs ($E_i$, $E_j$) for which $E_i^* = \emptyset$ and $E_j^* = \emptyset$ (i.e., no proxy event exceeds the similarity threshold $\tau$) cannot be grounded in the reference database and are therefore excluded from evaluation. Finally, we apply causal pair validation (§\ref{sec:causal_pair_validation}) for each pair, classifying it as \emph{valid}, \emph{invalid}, or \emph{not verifiable}.

\paragraph{Results} We report \emph{accuracy}, defined as the proportion of verifiable causal pairs that are supported by evidence in the reference database. Table \ref{tab:evaluation_validity} compares graph construction methods instantiated with the same LLM, thereby isolating the effect of the construction framework. Across methods, a substantial proportion of verifiable causal relations is valid. In the WoC methods, accuracy remains more stable with graph construction via chain discovery than with end-to-end graph construction. This is demonstrated in the results for PolarIs4CAUS, for which we have a higher number of verifiable causal pairs.

\subsection{Beyond Binary Validity Evaluation} \label{sec:important_intermediate_events}

Using a reference database of scientifically validated causal relations as the basis for evaluation offers clear advantages. Because it codifies relations as modular units, it can be reused in current and future climate change mining tasks. Moreover, grounding evaluation in vetted scientific statements ensures that performance measurements reflect alignment with established scientific knowledge. 

Our graph validity evaluation therefore primarily focused on micro-level pairwise relations. This choice is practical and aligns with how the reference database was constructed. Nevertheless, database-based evaluation can, in principle, extend beyond pairwise validation. One could assess whether graph construction methods capture key intermediate events within the generated causal graphs. Here, key events are those events that lie on the path(s) linking $E_c$ to $E_f$ in the reference database. The shift in our graph construction methods from pairwise relation verification toward sequence-level reasoning, where the graph grows conditioned on the context of a partially realized causal chain at each prediction step in the iterative approaches, suggests a natural extension toward evaluating sequence-level correctness. However, reliably and quantitatively evaluating these higher-level aspects is not feasible given the restricted size and coverage of climate change causality in the IPCC reports. With respect to key event inclusion, the two datasets contain too few verifiable causal pairs $E_c \rightarrow E_f$  with intermediate events (none in PolarIs3CAUS and four in PolarIs4CAUS) to meaningfully analyze this ability here. Sequence-level correctness is similarly difficult to assess: causal chains containing unverifiable relations (i.e., relations without corresponding entries in the database) may be wrongly interpreted as erroneous. Broadening coverage will likely require a database far larger than the one manually curated for this study, potentially created by extracting relations from large scientific corpora. 

\section{Conclusion}

This work introduced implicit causal graph construction, for which it designed and compared three construction approaches revealing the \textit{implicit causal mechanisms} explaining causality between cause and effect in text. Follow-up work is strongly encouraged to examine the broader applications of implicit causal graph construction to support the systematic analysis of latent causal beliefs and reasoning pathways, with direct implications for polarization studies and fact checking.

\section*{Acknowledgments} This work was funded by the Research Foundation - Flanders (FWO) under grant G0L0822N. Liesbeth Allein is further supported by a junior postdoctoral fellowship from the FWO under grant 12AGW26N.

\bibliographystyle{named}
\bibliography{ijcai2026}

\begin{thebibliography}{}

\bibitem[\protect\citeauthoryear{Ahn and Bailenson}{1996}]{ahn1996mechanism}
Woo-kyoung Ahn and Jeremy Bailenson.
\newblock {Causal Attribution as a Search for Underlying Mechanisms: An Explanation of the Conjunction Fallacy and the Discounting Principle}.
\newblock {\em Cognitive Psychology}, 31:82--123, 1996.

\bibitem[\protect\citeauthoryear{Allein \bgroup \em et al.\egroup }{2025}]{allein2025assessing}
Liesbeth Allein, Nataly Pineda-Casta{\~n}eda, Andrea Rocci, and Marie-Francine Moens.
\newblock {Assessing LLM Reasoning Through Implicit Causal Chain Discovery in Climate Discourse}.
\newblock {\em arXiv preprint arXiv:2510.13417}, 2025.

\bibitem[\protect\citeauthoryear{Allein \bgroup \em et al.\egroup }{2026}]{allein2026climatecause}
Liesbeth Allein, Nataly Pineda-Casta{\~n}eda, Andrea Rocci, and Marie-Francine Moens.
\newblock {ClimateCause: Complex and Implicit Causal Structures in Climate Reports}.
\newblock {\em arXiv preprint arXiv:2604.14856}, 2026.

\bibitem[\protect\citeauthoryear{Antonucci \bgroup \em et al.\egroup }{2023}]{antonucci2023zero}
Alessandro Antonucci, Gregorio Piqu{\'e}, and Marco Zaffalon.
\newblock {Zero-shot Causal Graph Extrapolation from Text via LLMs}.
\newblock {\em arXiv preprint arXiv:2312.14670}, 2023.

\bibitem[\protect\citeauthoryear{Chi \bgroup \em et al.\egroup }{2024}]{chi2024unveiling}
Haoang Chi, He~Li, Wenjing Yang, Feng Liu, Long Lan, Xiaoguang Ren, Tongliang Liu, and Bo~Han.
\newblock {Unveiling Causal Reasoning in Large Language Models: Reality or Mirage?}
\newblock {\em Advances in Neural Information Processing Systems}, 37:96640--96670, 2024.

\bibitem[\protect\citeauthoryear{Feldman and Hart}{2018}]{feldman2018climate}
Lauren Feldman and P~Sol Hart.
\newblock {Climate change as a polarizing cue: Framing effects on public support for low-carbon energy policies}.
\newblock {\em Global Environmental Change}, 51:54--66, 2018.

\bibitem[\protect\citeauthoryear{Feng \bgroup \em et al.\egroup }{2025}]{feng-etal-2025-iris}
Tao Feng, Lizhen Qu, Niket Tandon, and Gholamreza Haffari.
\newblock {{IRIS}: An Iterative and Integrated Framework for Verifiable Causal Discovery in the Absence of Tabular Data}.
\newblock In Wanxiang Che, Joyce Nabende, Ekaterina Shutova, and Mohammad~Taher Pilehvar, editors, {\em Proceedings of the 63rd Annual Meeting of the Association for Computational Linguistics (Volume 1: Long Papers)}, pages 9400--9428, Vienna, Austria, July 2025. Association for Computational Linguistics.

\bibitem[\protect\citeauthoryear{Heindorf \bgroup \em et al.\egroup }{2020}]{heindorf2020causenet}
Stefan Heindorf, Yan Scholten, Henning Wachsmuth, Axel-Cyrille Ngonga~Ngomo, and Martin Potthast.
\newblock {CauseNet: Towards a Causality Graph Extracted from the Web}.
\newblock In {\em {Proceedings of the 29th ACM International Conference on Information \& Knowledge Management}}, pages 3023--3030, 2020.

\bibitem[\protect\citeauthoryear{Hwang \bgroup \em et al.\egroup }{2021}]{hwang2021comet}
Jena~D Hwang, Chandra Bhagavatula, Ronan Le~Bras, Jeff Da, Keisuke Sakaguchi, Antoine Bosselut, and Yejin Choi.
\newblock {(Comet-) Atomic 2020: On Symbolic and Neural Commonsense Knowledge Graphs}.
\newblock In {\em {Proceedings of the AAAI Conference on Artificial Intelligence}}, volume~35, pages 6384--6392, 2021.

\bibitem[\protect\citeauthoryear{Jaimini and Sheth}{2022}]{9706608}
Utkarshani Jaimini and Amit Sheth.
\newblock {CausalKG: Causal Knowledge Graph Explainability Using Interventional and Counterfactual Reasoning}.
\newblock {\em {IEEE Internet Computing}}, 26(1):43--50, 2022.

\bibitem[\protect\citeauthoryear{Jiralerspong \bgroup \em et al.\egroup }{2024}]{jiralerspong2024efficient}
Thomas Jiralerspong, Xiaoyin Chen, Yash More, Vedant Shah, and Yoshua Bengio.
\newblock {Efficient Causal Graph Discovery Using Large Language Models}.
\newblock {\em arXiv preprint arXiv:2402.01207}, 2024.

\bibitem[\protect\citeauthoryear{Kiciman \bgroup \em et al.\egroup }{2023}]{kiciman2023causal}
Emre Kiciman, Robert Ness, Amit Sharma, and Chenhao Tan.
\newblock {Causal Reasoning and Large Language Models: Opening a New Frontier for Causality}.
\newblock {\em {Transactions on Machine Learning Research}}, 2023.

\bibitem[\protect\citeauthoryear{Kocsis and Szepesv{\'a}ri}{2006}]{kocsis2006bandit}
Levente Kocsis and Csaba Szepesv{\'a}ri.
\newblock {Bandit Based Monte-Carlo Planning}.
\newblock In {\em {European Conference on Machine Learning}}, pages 282--293. Springer, 2006.

\bibitem[\protect\citeauthoryear{Koupaee \bgroup \em et al.\egroup }{2025}]{koupaee-etal-2025-causal}
Mahnaz Koupaee, Xueying Bai, Mudan Chen, Greg Durrett, Nathanael Chambers, and Niranjan Balasubramanian.
\newblock {Causal Graph based Event Reasoning using Semantic Relation Experts}.
\newblock In Wanxiang Che, Joyce Nabende, Ekaterina Shutova, and Mohammad~Taher Pilehvar, editors, {\em Proceedings of the 63rd Annual Meeting of the Association for Computational Linguistics (Volume 1: Long Papers)}, pages 26169--26199, Vienna, Austria, July 2025. Association for Computational Linguistics.

\bibitem[\protect\citeauthoryear{Li \bgroup \em et al.\egroup }{2021}]{10.5555/3491440.3491942}
Zhongyang Li, Xiao Ding, Ting Liu, J.~Edward Hu, and Benjamin Van~Durme.
\newblock {Guided Generation of Cause and Effect}.
\newblock In {\em Proceedings of the Twenty-Ninth International Joint Conference on Artificial Intelligence}, IJCAI'20, pages 3629--3636, 2021.

\bibitem[\protect\citeauthoryear{Liu \bgroup \em et al.\egroup }{2024}]{liu-etal-2024-identifying}
Cheng Liu, Wei Xiang, and Bang Wang.
\newblock {Identifying while Learning for Document Event Causality Identification}.
\newblock In Lun-Wei Ku, Andre Martins, and Vivek Srikumar, editors, {\em Proceedings of the 62nd Annual Meeting of the Association for Computational Linguistics (Volume 1: Long Papers)}, pages 3815--3827, Bangkok, Thailand, August 2024. Association for Computational Linguistics.

\bibitem[\protect\citeauthoryear{Long \bgroup \em et al.\egroup }{2023}]{long2023causal}
Stephanie Long, Alexandre Pich{\'e}, Valentina Zantedeschi, Tibor Schuster, and Alexandre Drouin.
\newblock {Causal Discovery with Language Models as Imperfect Experts}.
\newblock In {\em ICML 2023 Workshop on Structured Probabilistic Inference {\&} Generative Modeling}, 2023.

\bibitem[\protect\citeauthoryear{Minnerop and Otto}{2019}]{minnerop2019climate}
Petra Minnerop and Friederike Otto.
\newblock {Climate Change and Causation: Joining Law and Climate Science on the Basis of Formal Logic}.
\newblock {\em Buff. Envtl. LJ}, 27:49, 2019.

\bibitem[\protect\citeauthoryear{Nicholson}{2014}]{nicholson2014climate}
Calum~TM Nicholson.
\newblock Climate change and the politics of causal reasoning: the case of climate change and migration.
\newblock {\em The Geographical Journal}, 180(2):151--160, 2014.

\bibitem[\protect\citeauthoryear{Pearl}{2009}]{pearl2009causality}
Judea Pearl.
\newblock {\em Causality}.
\newblock Cambridge university press, 2009.

\bibitem[\protect\citeauthoryear{Pineda and Allein}{2025a}]{pineda2025polaris3}
Nataly Pineda and Liesbeth Allein.
\newblock {PolarIs3CAUS (Version 1.0) [Dataset]}, 2025.

\bibitem[\protect\citeauthoryear{Pineda and Allein}{2025b}]{pineda2025polaris4}
Nataly Pineda and Liesbeth Allein.
\newblock {PolarIs4CAUS (Version 1.0) [Dataset]}, 2025.

\bibitem[\protect\citeauthoryear{Sap \bgroup \em et al.\egroup }{2019}]{sap2019atomic}
Maarten Sap, Ronan Le~Bras, Emily Allaway, Chandra Bhagavatula, Nicholas Lourie, Hannah Rashkin, Brendan Roof, Noah~A Smith, and Yejin Choi.
\newblock {ATOMIC: An Atlas of Machine Commonsense for If-Then Reasoning}.
\newblock In {\em {Proceedings of the AAAI Conference on Artificial Intelligence}}, volume~33, pages 3027--3035, 2019.

\bibitem[\protect\citeauthoryear{Speer \bgroup \em et al.\egroup }{2017}]{speer2017conceptnet}
Robyn Speer, Joshua Chin, and Catherine Havasi.
\newblock {ConceptNet 5.5: An Open Multilingual Graph of General Knowledge}.
\newblock In {\em {Proceedings of the AAAI Conference on Artificial Intelligence}}, volume~31, 2017.

\bibitem[\protect\citeauthoryear{Tan \bgroup \em et al.\egroup }{2023}]{tan-etal-2023-recess}
Fiona~Anting Tan, Hansi Hettiarachchi, Ali H{\"u}rriyeto{\u{g}}lu, Nelleke Oostdijk, Tommaso Caselli, Tadashi Nomoto, Onur Uca, Farhana~Ferdousi Liza, and See-Kiong Ng.
\newblock {{RECESS}: Resource for Extracting Cause, Effect, and Signal Spans}.
\newblock In Jong~C. Park, Yuki Arase, Baotian Hu, Wei Lu, Derry Wijaya, Ayu Purwarianti, and Adila~Alfa Krisnadhi, editors, {\em Proceedings of the 13th International Joint Conference on Natural Language Processing and the 3rd Conference of the Asia-Pacific Chapter of the Association for Computational Linguistics (Volume 1: Long Papers)}, pages 66--82, Nusa Dua, Bali, November 2023. Association for Computational Linguistics.

\bibitem[\protect\citeauthoryear{Vashishtha \bgroup \em et al.\egroup }{2023}]{vashishtha2023causal}
Aniket Vashishtha, Abbavaram~Gowtham Reddy, Abhinav Kumar, Saketh Bachu, Vineeth~N Balasubramanian, and Amit Sharma.
\newblock {Causal Inference using LLM-Guided Discovery}.
\newblock In {\em AAAI 2024 Workshop on ``Are Large Language Models Simply Causal Parrots?''}, 2023.

\bibitem[\protect\citeauthoryear{Vo \bgroup \em et al.\egroup }{2025}]{vo-etal-2025-access}
Vy~Vo, Lizhen Qu, Tao Feng, Yuncheng Hua, Xiaoxi Kang, Songhai Fan, Tim Dwyer, Lay-Ki Soon, and Gholamreza Haffari.
\newblock {{ACCESS} : A Benchmark for Abstract Causal Event Discovery and Reasoning}.
\newblock In Luis Chiruzzo, Alan Ritter, and Lu~Wang, editors, {\em Proceedings of the 2025 Conference of the Nations of the Americas Chapter of the Association for Computational Linguistics: Human Language Technologies (Volume 1: Long Papers)}, pages 1049--1074, Albuquerque, New Mexico, April 2025. Association for Computational Linguistics.

\bibitem[\protect\citeauthoryear{Walsh and Sloman}{2011}]{walsh2011meaning}
Clare~R Walsh and Steven~A Sloman.
\newblock {The Meaning of Cause and Prevent: The Role of Causal Mechanism}.
\newblock {\em Mind \& Language}, 26(1):21--52, 2011.

\bibitem[\protect\citeauthoryear{Wang}{2024}]{wang-2024-causalbench}
Zeyu Wang.
\newblock {{C}ausal{B}ench: A Comprehensive Benchmark for Evaluating Causal Reasoning Capabilities of Large Language Models}.
\newblock In Kam-Fai Wong, Min Zhang, Ruifeng Xu, Jing Li, Zhongyu Wei, Lin Gui, Bin Liang, and Runcong Zhao, editors, {\em Proceedings of the 10th SIGHAN Workshop on Chinese Language Processing (SIGHAN-10)}, pages 143--151, Bangkok, Thailand, August 2024. Association for Computational Linguistics.

\bibitem[\protect\citeauthoryear{Xiong \bgroup \em et al.\egroup }{2025}]{xiong-etal-2025-com2}
Kai Xiong, Xiao Ding, Yixin Cao, Yuxiong Yan, Li~Du, Yufei Zhang, Jinglong Gao, Jiaqian Liu, Bing Qin, and Ting Liu.
\newblock {Com$^2$ : A Causal-Guided Benchmark for Exploring Complex Commonsense Reasoning in Large Language Models}.
\newblock In Wanxiang Che, Joyce Nabende, Ekaterina Shutova, and Mohammad~Taher Pilehvar, editors, {\em Proceedings of the 63rd Annual Meeting of the Association for Computational Linguistics (Volume 1: Long Papers)}, pages 16119--16140, Vienna, Austria, July 2025. Association for Computational Linguistics.

\end{thebibliography}

\appendix

\section{Reproducibility}

\subsection{Data Availability}

The experiments in this work rely on two existing datasets with cause-effect pairs extracted from text, i.e., PolarIs3CAUS and PolarIs4CAUS. Causal validity of causal pairs in the predicted graphs is evaluated using a novel database with scientifically validated cause-effect pairs, which is specifically curated for this purpose.

\paragraph{Existing datasets} While the datasets are properly cited in the main paper, the authors are still in the process of formally publishing them in their national research database, SWISSUbase. In the meantime, limited but detailed documentation, including a codebook that clearly describes the annotation guidelines and dataset structure, is already publicly available for PolarIs4CAUS at the following URL: \url{https://doi.org/10.48656/g07p-nq64}. PolarIs3CAUS follows a similar annotation scheme and structure. However, corresponding documentation is not yet accessible through SWUSSUbase. Nevertheless, both datasets have been made publicly available via a GitHub repository associated with another, currently non‑archival paper: \url{https://github.com/laallein/implicit-causal-chain-discovery}. Two JSON files with the causal pairs from the datasets which are used in the experiments (excluding the statements from which they are extracted and other metadata) are included in the supplementary materials (in the Code folder): \textit{`qa\_polaris\_3.json'} (PolarIs3CAUS) and \textit{`qa\_polaris\_4.json'} (PolarIs4CAUS).

The use of the two datasets in this work complies with their usage license (NCCR Evolving Language – Restricted access: for research only).

\paragraph{Novel database} Two JSON files, one containing the unique events (\textit{`REFERENCE\_DATABASE\_events.json'}) and the other containing all causal pairs (\textit{`REFERENCE\_DATABASE\_relations.json'}), together make up the ground-truth reference database. These files will be made publicly available in a GitHub repository upon publication of the paper. 

\subsection{Computational Experiments}

\paragraph{Code} We include all code required for conducting the experiments will make it publicly available in a GitHub repository upon publication of the paper.

\paragraph{Implementations} We rely on the NetworkX package in Python for representing graphs and performing actions over a graph, e.g. studying their structure: \url{https://networkx.org/documentation/stable/index.html}. For performing fuzzy matching, we rely on the Rapidfuzz package: \url{https://rapidfuzz.github.io/RapidFuzz/Usage/fuzz.html\#ratio}

Model implementations:
\begin{itemize}
    \item GPT4o-mini: gpt-4o-mini-2024-07-18 (OpenAI API)
    \item Falcon3-7b: tiiuae/falcon3-7b-instruct (NVIDIA NIM)
    \item Qwen2-7b: qwen/qwen2-7b-instruct (NVIDIA NIM)
    \item Llama 3.1 Nemotron: nvidia/llama-3.1-nemotron-ultra-253b-v1 (NVIDIA NIM)
    \item DeepSeek R1: deepseek-ai/deepseek-r1 (NVIDIA NIM)
    \item Mixtral: mistralai/mixtral-8x22b-instruct-v0.1 (NVIDIA NIM)
    \item Phi 4-mini: microsoft/phi-4-mini-instruct (NVIDIA NIM)
    \item Mistral Nemo: nv-mistralai/mistral-nemo-12b-instruct (NVIDIA NIM)
    \item Llama 3 70b: meta/llama3-70b-instruct (NVIDIA NIM)
    \item GPT4o: gpt-4o-2024-11-20 (OpenAI API)
    \item o1: o1-preview-2024-09-12 (OpenAI API)
    \item o1-mini: o1-mini-2024-09-12 (OpenAI API)
\end{itemize}

\paragraph{Hyperparameters} Table \ref{tab:hyperparameters} outlines hyperparameters, with for some the number and range of values tried  during the development of the experiments in this paper. Final values are underlined. Unless stated otherwise, we rely on the standard hyperparameters set by the developers of the LLMs for the LLM-related hyperparameters.

Note on rollout. When including the generated nodes from the rollout in the graph, the value of $V(n)$ of the generated nodes would be set to $0$, which causes an error in the UCT computation as it would cause divisions by zero. Therefore, we set the UCT to $+\infty$ for nodes with $V(n) = 0$. 

\begin{table*}[]
    \centering
    \footnotesize
    \begin{tabular}{p{4cm}p{4cm}p{7cm}}
    \toprule
       \textbf{Hyperparameters} & \textbf{Values tried} & \textbf{Motivation} \\
    \midrule
    Exploration weight $\lambda$ & $2.0$ & \\ 
    Number of children $J$ & \{$2$, \underline{$3$}, $5$\} & $2$ was too small to capture enough interesting events while $5$ was too large, harming the validity of the next predicted causal events. \\
    Guidance prompt $g$ & \{\textit{\underline{``this leads to"}}, \textit{``causes''}\} & The guidance phrase \textit{``this leads to''} is chosen over alternatives like \textit{``causes''} to emphasize the effect to be generated, rather than the initiating cause. \\
    Fuzzy matching threshold $\tau$ & $\{.70, \underline{.75}, .80, .85, .90\}$ & We manually verified the relevance of retrieved events across different threshold values, applying. The final value was chosen to balance semantic similarity with the number of retained proxy events. \\
    Number of MCTS iterations & \{\underline{10}, \underline{20}, 50, 100\} & A higher number of MCTS iterations did not result in higher-quality graphs. \\
    Maximum rollout length & $8$ & \\
    \midrule
    \multicolumn{3}{c}{LLM-related hyperparameters (those implemented through NVIDIA NIM)} \\
    \midrule
    Temperature & $0.6$ & \\
    Top-p & $0.6$ & \\
    Max tokens & $4096$ \\
    Seed & $256$ & \\
    \bottomrule
    \end{tabular}
    \caption{Overview of the hyperparameters.}
    \label{tab:hyperparameters}
\end{table*}

\section{Overview of the LLM Prompts}

\paragraph{End-to-end graph generation} \textit{``You are tasked with mapping a given initial cause and final effect to a complete causal graph. A causal graph is a directed acyclic graph where nodes represent noun phrase formulations of events and directed edges represent the causal relation from one event to another. Rules: 1. The causal graph has one source node that contains the given initial cause and one sink node that contains the given final effect. 2. All intermediate nodes in the graph should be predicted by you and the nodes are only formulated as noun phrases. 3. You can predict as many intermediate nodes as needed to have a causal graph that faithfully contains all possible intermediate events explaining the causal mechanisms linking the initial cause to the final effect. 4. Be as exhaustive as possible. 5. All nodes should be part of a path linking the initial cause to the final effect. Format the causal graph as an adjacency list: (initial cause, event\_1) (initial cause, event\_2) (event\_1, event\_3) (event\_2, event\_4) ... (event\_N, final\_effect). Initial cause: \{$E_c$\}. Final effect: \{$E_f$\}. Causal graph:"}

\paragraph{Graph generation via causal chain discovery (single pass)} \textit{``A causal chain is a sequence of events in which each event directly causes the next, forming a connected series of cause-and-effect relations. Unfolding a causal chain means identifying and linking individual events. A step of the chain presents only one noun phrase containing the event. Unfold all possible causal chains that connect \texttt{\{$E_c$\}} (initial cause) to \texttt{\{$E_f$\}} (final effect) and separate the steps of the chain with the token \textless step\textgreater, and the chains with the token \textless chain\textgreater."}

\paragraph{Graph generation via causal chain discovery (iterative)} \textbf{System prompt}: \textit{``You are tasked with unfolding a causal chain that begins with a given initial cause and ends with a final effect. A causal chain is a sequence of events, where each event directly leads to the next, forming a clear cause-and-effect progression. You build the chain incrementally, identifying the next node based on the last node in the chain built thus far. You start with the initial cause and build towards the final effect. Guidelines: Represent an event with a single noun phrase. Each turn produces ONE event. NEVER identify multiple events at once, and NEVER explain or discuss your reasoning. When the event is the final effect, format your output as "end of chain: final effect"}

\textbf{Guidance prompt}: \textit{``this leads to''}.

\section{Detailed Breakdown of the Results}

Detailed breakdown of results for the causal graph construction methods are included in Table \ref{tab:evaluation_structural_properties_ETE}, \ref{tab:evaluation_structural_properties_SP}, and \ref{tab:evaluation_structural_properties_mcts}.

\begin{table}[]
    \centering
    \tiny
    \setlength{\tabcolsep}{1pt}
    \begin{tabular}{p{1.3cm}|p{0.8cm}p{0.8cm}p{0.8cm}p{0.6cm}|p{0.8cm}p{0.8cm}p{0.8cm}p{0.6cm}}
        \toprule
        & \multicolumn{4}{c}{Polaris3CAUS} & \multicolumn{4}{c}{Polaris4CAUS} \\
        \midrule
        \textbf{Setup} 
        & \textbf{Graph Density ($\uparrow$)} 
        & \textbf{Longest Path Length ($\uparrow$)} 
        & \textbf{Avg. Node Connectivity ($\uparrow$)} 
        & \textbf{Cyclic Ratio ($\downarrow$)} 
        & \textbf{Graph Density ($\uparrow$)} 
        & \textbf{Longest Path Length ($\uparrow$)} 
        & \textbf{Avg. Node Connectivity ($\uparrow$)} 
        & \textbf{Cyclic Ratio ($\downarrow$)} \\
        \midrule

        \multicolumn{9}{c}{\textit{\textbf{End-to-End} Causal Graph Generation}} \\
        \midrule
        GPT4o-mini
        & 0.11 (±0.04) & 6.70 (±2.43) & 0.39 (±0.11) & 0.02
        & 0.12 (±0.04) & 6.50 (±2.24) & 0.40 (±0.10) & 0.08 \\
        Falcon3
        & 0.33 (±0.18) & 2.66 (±2.62) & 0.44 (±0.12) & 0.02
        & 0.31 (±0.17) & 3.03 (±2.07) & 0.45 (±0.09) & 0.00  \\
        Qwen2 
        & 0.43 (±0.12) & 1.10 (±0.30) & 0.44 (±0.11) & 0.01
        & 0.45 (±0.11) & 1.05 (±0.22) & 0.45 (±0.11) & 0.00  \\
        \textit{Single Model*} 
        & \textit{0.29 (±0.11)} & \textit{3.49 (±1.78)} & \textit{0.42 (±0.11)} & \textit{0.01}
        & \textit{0.29 (±0.11)} &\textit{ 3.53 (±1.51)} & \textit{0.44 (±0.10)} & \textit{0.02}  \\
        \midrule
        \textit{Post-hoc WoC} 
        & 0.16 (±0.17) & 5.18 (±3.23) & 0.29 (±0.16) & 0.02 
        & 0.14 (±0.16) & 5.55 (±2.80) & 0.27 (±0.14) & 0.05 \\
        \bottomrule
    \end{tabular}
    \caption{End-to-end causal graph generation: structural graph evaluation.}
    \label{tab:evaluation_structural_properties_ETE}
\end{table}

\begin{table}[]
    \centering
    \tiny
    \setlength{\tabcolsep}{1pt}
    \begin{tabular}{p{1.3cm}|p{0.8cm}p{0.8cm}p{0.8cm}p{0.6cm}|p{0.8cm}p{0.8cm}p{0.8cm}p{0.6cm}}
        \toprule
        & \multicolumn{4}{c}{Polaris3CAUS} & \multicolumn{4}{c}{Polaris4CAUS} \\
        \midrule
        \textbf{Setup} 
        & \textbf{Graph Density ($\uparrow$)} 
        & \textbf{Longest Path Length ($\uparrow$)} 
        & \textbf{Avg. Node Connectivity ($\uparrow$)} 
        & \textbf{Cyclic Ratio ($\downarrow$)} 
        & \textbf{Graph Density ($\uparrow$)} 
        & \textbf{Longest Path Length ($\uparrow$)} 
        & \textbf{Avg. Node Connectivity ($\uparrow$)} 
        & \textbf{Cyclic Ratio ($\downarrow$)} \\
        \midrule
        \multicolumn{9}{c}{Via \textit{\textbf{Single-Pass} Causal Chain Discovery}} \\
        \midrule
        o1 & 0.06 (±0.04) & 5.26 (±1.91) & 0.16 (±0.09) & 0.02 & 0.06 (±0.03) & 5.34 (±1.23) & 0.15 (±0.07) & 0.01 \\
    o1-mini & 0.08 (±0.04) & 4.28 (±1.02) & 0.19 (±0.07) & 0.01 & 0.08 (±0.03) & 4.32 (±3.07) & 0.19 (±0.06) & 0.01 \\
    GPT4o & 0.08 (±0.05) & 5.51 (±1.95) & 0.23 (±0.10) & 0.02 & 0.09 (±0.04) & 5.05 (±1.53) & 0.24 (±0.09) & 0.02 \\
    DeepSeek R1 & 0.04 (±0.01) & 5.88 (±1.74) & 0.13 (±0.04) & 0.00 & 0.05 (±0.02) & 5.74 (±1.37) & 0.14 (±0.04) & 0.00 \\
    Llama 3.1 Nemo & 0.07 (±0.03) & 6.38 (±2.19) & 0.24 (±0.09) & 0.02 & 0.08 (±0.03) & 6.37 (±2.06) & 0.26 (±0.10) & 0.03 \\
    Llama 3 70b & 0.08 (±0.03) & 5.65 (±1.62) & 0.24 (±0.09) & 0.01 & 0.08 (±0.03) & 5.33 (±1.27) & 0.23 (±0.07) & 0.01 \\
    Mistral Nemo & 0.06 (±0.05) & 4.96 (±1.33) & 0.15 (±0.09) & 0.02 & 0.04 (±0.02) & 6.53 (±2.16) & 0.13 (±0.07) & 0.02 \\
    Mixtral & 0.07 (±0.03) & 6.30 (±2.09) & 0.24 (±0.11) & 0.07 & 0.07 (±0.03) & 6.16 (±1.75) & 0.25 (±0.10) & 0.02 \\
    Phi 4-mini & 0.09 (±0.06) & 7.00 (±3.08) & 0.32 (±0.15) & 0.06 & 0.08 (±0.05) & 7.60 (±4.27) & 0.30 (±0.14) & 0.09 \\
    \midrule
    \textit{Single Model}** 
        & \textit{0.07 (±0.04)} & \textit{5.69 (±1.88)} & \textit{0.21 (±0.09)} & \textit{0.02}
        & \textit{0.07 (±0.03}) & \textit{5.83 (±2.08)} & \textit{0.21 (±0.08)} & \textit{0.02} \\
    \midrule
    \textit{Only general-purpose LMs}: \{GPT4o, Mixtral, Phi 4-mini\} & 0.03 (±0.01) & 8.18 (±3.35) & 0.10 (±0.05) & 0.16 & 0.03 (±0.01) & 8.76 (±4.07) & 0.11 (±0.05) & 0.15 \\
    \textit{Only reasoning LMs}: \{o1, o1-mini, DeepSeek R1\} & 0.03 (±0.01) & 5.95 (±1.67) & 0.08 (±0.02) & 0.11 & 0.03 (±0.01) & 6.08 (±3.13) & 0.09 (±0.03) & 0.02 \\
    \textit{Mixed}: \{o1, Mixtral, DeepSeek R1\} & 0.02 (±0.01) & 7.02 (±2.17) & 0.06 (±0.02) & 0.08 & 0.02 (±0.01) & 6.85 (±1.71) & 0.07 (±0.02) & 0.03 \\
    \midrule
        \textit{Post-hoc WoC}* (3 LLMs) 
        & \textit{0.03 (±0.01}) & \textit{7.05 (±2.40)} & \textit{0.08 (±0.03)} & \textit{0.08}
        & \textit{0.03 (±0.01)} & \textit{7.23 (±2.97)} & \textit{0.09 (±0.04)} & \textit{0.07} \\
    \midrule
        \textit{Post-hoc WoC} (9 LLMs)
        & \textit{0.01 (±0.00)} & \textit{9.29 (±2.88)} & \textit{0.03 (±0.01)} & \textit{0.24}
        & \textit{0.01 (±0.00)} & \textit{10.41 (±4.04)} & \textit{0.03 (±0.02)} &\textit{ 0.30} \\

        \bottomrule
    \end{tabular}
    \caption{Via single-pass causal chain discovery: structural graph evaluation.}
    \label{tab:evaluation_structural_properties_SP}
\end{table}

\begin{table}[]
    \centering
    \tiny
    \setlength{\tabcolsep}{1pt}
    \begin{tabular}{p{1.3cm}|p{0.8cm}p{0.8cm}p{0.8cm}p{0.6cm}|p{0.8cm}p{0.8cm}p{0.8cm}p{0.6cm}}
        \toprule
        & \multicolumn{4}{c}{Polaris3CAUS} & \multicolumn{4}{c}{Polaris4CAUS} \\
        \midrule
        \textbf{Setup} 
        & \textbf{Graph Density ($\uparrow$)} 
        & \textbf{Longest Path Length ($\uparrow$)} 
        & \textbf{Avg. Node Connectivity ($\uparrow$)} 
        & \textbf{Cyclic Ratio ($\downarrow$)} 
        & \textbf{Graph Density ($\uparrow$)} 
        & \textbf{Longest Path Length ($\uparrow$)} 
        & \textbf{Avg. Node Connectivity ($\uparrow$)} 
        & \textbf{Cyclic Ratio ($\downarrow$)} \\
        \midrule
        \multicolumn{9}{c}{\textit{Via \textbf{Iterative} Causal Chain Discovery}} \\
        \midrule

        GPT4o (\textit{10 it.}) 
        & 0.09 (±0.03) & 7.17 (±1.76) & 0.30 (±0.10) & -
        & 0.09 (±0.04) & 7.51 (±1.70) & 0.31 (±0.11) & - \\

        Qwen2 (\textit{10 it.}) 
        & 0.32 (±0.16) & 3.00 (±2.04) & 0.5 (±0.00) & -
        & 0.32 (±0.15) & 2.95 (±1.83) & 0.5 (±0.00) & - \\

        Falcon3 (\textit{10 it.}) 
        & 0.36 (±0.10) & 2.00 (±0.87) & 0.50 (±0.00) & -
        & 0.35 (±0.11) & 2.19 (±1.15) & 0.5 (±0.00) & - \\

        \midrule
        \textit{Single Model} (\textit{10 it.}) 
        & 0.26 (±0.10) & 4.06 (±1.55) & 0.43 (±0.03) & -
        & 0.25 (±0.10) & 4.22 (±1.56) & 0.44 (±0.04) & - \\
        \midrule

        GPT4o (\textit{20 it.}) 
        & 0.07 (±0.03) & 9.06 (±3.05) & 0.30 (±0.09) & -
        & 0.07 (±0.03) & 8.41 (±2.10) & 0.31 (±0.10) & - \\

        Qwen2 (\textit{20 it.}) 
        & 0.32 (±0.16) & 3.00 (±2.04) & 0.5 (±0.00) & -
        & 0.32 (±0.15) & 2.95 (±1.83) & 0.5 (±0.00) & - \\

        Falcon3 (\textit{20 it.}) 
        & 0.36 (±0.10) & 2.00 (±0.87) & 0.50 (±0.00) & -
        & 0.35 (±0.11) & 2.18 (±1.14) & 0.50 (±0.00) & - \\
        \midrule

        \textit{Single Model} (\textit{20 it.}) 
        & 0.25 (±0.10) & 4.69 (±1.98) & 0.43 (±0.04) & -
        & 0.25 (±0.10) & 4.52 (±1.70) & 0.44 (±0.03) & - \\
        \midrule
        \midrule

        \textit{Post-hoc WoC} (\textit{10 it.}) 
        & 0.10 (±0.08) & 6.96 (±2.02) & 0.30 (±0.11) & -
        & 0.11 (±0.11) & 6.96 (±2.35) & 0.32 (±0.12) & - \\

        \textit{Post-hoc WoC} (\textit{20 it.}) 
        & 0.08 (±0.08) & 8.62 (±3.24) & 0.29 (±0.10) & -
        & 0.10 (±0.10) & 7.73 (±2.78) & 0.31 (±0.11) & - \\

        \midrule
        \midrule

        \textit{Collab. WoC} (\textit{10 it.})  
        & 0.13 (±0.10) & 5.53 (±2.77) & 0.27 (±0.12) & -
        & 0.13 (±0.09) & 5.48 (±2.69) & 0.29 (±0.10) & - \\

        \textit{Collab. WoC} (\textit{20 it.}) 
        & 0.19 (±0.13) & 4.07 (±3.51) & 0.29 (±0.11) & -
        & 0.14 (±0.09) & 5.18 (±2.84) & 0.27 (±0.10) & - \\

        \bottomrule
    \end{tabular}
    \caption{Via iterative causal chain discovery: structural graph evaluation.}
    \label{tab:evaluation_structural_properties_mcts}
\end{table}

\begin{table}[t!]
\centering
\scriptsize
\begin{tabular}{p{3.1cm}p{1.75cm}p{1.75cm}}
\textbf{} & \textbf{PolarIs3CAUS} & \textbf{PolarIs4CAUS} \\
\hline
Total \# causal pairs & 95  & 181  \\
$\rightarrow$ Matches for cause only & 15 \textit{(.16)} & 38 \textit{(.21)} \\
$\rightarrow$ Matches for effect only & 13 \textit{(.14)} & 32 \textit{(.18)} \\
$\rightarrow$ Matches for cause and effect & 2 \textit{(.02)} & 20 \textit{(.11)} \\
$\quad\;\rightarrow$ No path / \textbf{valid} / \textbf{invalid} & 2 / \textbf{0} / \textbf{0} & 12 / \textbf{8} / \textbf{0} \\
$\quad\;\rightarrow$ If \textbf{valid}: \textbf{with} / without intermediate node in reference & \textbf{0} / 0 & \textbf{4} / 4 \\
\hline
\end{tabular}
\caption{Validated causal pairs in PolarIs3CAUS and PolarIs4CAUS, with relative proportions in brackets.}
\label{tab:E1-results}
\end{table}

\section{Reference Database}

\subsection{Annotation Guidelines}

\paragraph{Cause-effect pair identification} A cause-effect pair, or causal pair, is a set of two events which are reported to have a causal relation in a given text, where the event acting as \textit{the cause} leads to the existence or change in the event, i.e., \textit{the effect}. Causal relations with the multi-component cause or effect (e.g., the cause encompasses more than one event) are broken down into their individual causal relations.

\paragraph{General rule for noun phrase reformulation} The cause and effect should be formulated as noun phrases in such a way that one can read the causal relation as \textit{``[Cause] causes/leads to [Effect]''}, and the causal relation can be understood as outside its communicative context. 

\paragraph{Reformulation guidelines} Make minimal alterations to the semantics of the cause and effect so that the reformulated version stays as close as possible to the semantics of the original phrasing. Rely on words used in the statement for reformulation as much as possible.
\\ \\
\textit{\textbf{Example:} ``With further global warming, every region is projected to increasingly experience concurrent and multiple changes in climatic impact-drivers.''}
\begin{itemize}
    \item Causal relation \#1: \textit{global warming} $\rightarrow$ \textit{increase in concurrent changes in climatic impact-drivers}
    \item Causal relation \#2: \textit{global warming} $\rightarrow$ \textit{increase in multiple changes in climatic impact-drivers}
\end{itemize}

\subsection{Samples from Reference Database}

Table \ref{tab:examples_database} present sample cause-effect pairs from the reference database.

\begin{table}[ht]
\footnotesize
\centering
\begin{tabular}{p{0.4\columnwidth} p{0.5\columnwidth}}
\hline
\textbf{Cause} & \textbf{Effect} \\
\hline
addressed governance constraints & soft limits reached by smallholder farmers \\
unbalanced diets & unsustainable agricultural expansion \\
integrated transport infrastructure operations & environmental impacts of decarbonising the transport sectors \\
relative sea level rise & risk for infrastructure \\
unsustainable land use & differences in vulnerability of people to climate change among regions \\
climate change & land degradation \\
unsustainable ocean use & differences in vulnerability of people to climate change within regions \\
marginalisation & differences in vulnerability of people to climate change among regions \\
inequity & differences in vulnerability of people to climate change among regions \\
human activities & stratospheric ozone depletion \\
net greenhouse gases emissions from production & human-caused climate change \\
urban form & urban greenhouse gases emissions \\
ocean acidification & food production from shellfish fisheries \\
integrated transport infrastructure planning & economic impacts of decarbonising the transport sectors \\
\hline
\end{tabular}
    \caption{Example cause-effect pairs from the reference database.}
    \label{tab:examples_database}
\end{table}

\section{Capturing Key Intermediate Events}

First, obtain the intermediate events in \emph{the generated graph} $G$ linking $E_c$ to $E_f$. Rather than considering all events in $G$, the analysis is restricted to events that lie on at least one directed path connecting the two endpoints. 
Let $Paths_{E_c \rightarrow E_f}$ be the set of paths from $E_c$ to $E_f$ in $G$, and $U_{gen}$ the set of unique intermediate events that occur on at least one path in $Paths_{E_c \rightarrow E_f}$, excluding $E_c$ and $E_f$. Obtain proxy events for each intermediate event in $U_{gen}$ through event matching and include them in the set $U^*_{gen}$. Second, obtain the intermediate events in \emph{the reference database}. Retrieve proxy event sets $E_c^*$ and $E_f^*$, and extract all paths from each $E_i \in E_c^*$ to any event $E_j \in E_f^*$, retaining them in path set $Paths_{E_c^* \rightarrow E_f^*}$. From $Paths_{E_c^* \rightarrow E_f^*}$, collect all unique events into the set $U^*_{ref}$. Finally, precision can be computed, reflecting the proportion of key events in the reference database captured by the model:
\begin{equation}
    \text{Precision} = \frac{|U^*_{ref} \cap U_{gen}^*|}{|U^*_{ref}|}
\end{equation}

\section{On the Use of AI Assistants in Coding and Writing}

In this research, artificial intelligence assistants were used to assist in coding and writing (Copilot). After using these tools/services, the authors reviewed and edited the content as needed and take full responsibility for the content of the publication.

\end{document}